\documentclass[sigconf]{acmart}
\AtBeginDocument{%
  }

\setcopyright{cc}
\setcctype{by}
\copyrightyear{2026}
\acmYear{2026}
\acmDOI{10.1145/3799682.3840809}
\acmConference[CIKM '26]
  {Proceedings of the 35th ACM International Conference on Information and Knowledge Management}
  {November 7--11, 2026}
  {Rome, Italy.}
\acmBooktitle{Proceedings of the 35th ACM International Conference on
  Information and Knowledge Management (CIKM '26), November 7--11, 2026,
  Rome, Italy}
\acmISBN{979-8-4007-2539-5/2026/11}
\usepackage{multirow}
\usepackage{enumitem}
\usepackage{makecell}
\usepackage{colortbl}

\begin{document}

\title{ClinSDT: LLM-Encoded Clinical Semantic Guidance for Sepsis Treatment via Decision Transformer}

\author{Yanyan Wu}
\authornote{Yanyan Wu and Guanyu Zhu contributed equally to this work.}
\email{23013032@mail.ecust.edu.cn}
\affiliation{%
  \institution{East China University of Science and Technology}
  \city{Shanghai}
  \country{China}}

\author{Guanyu Zhu}
\authornotemark[1]
\email{xiaoguanbx@gmail.com}
\affiliation{%
  \institution{South China Agricultural University}
  \city{Guangzhou}
  \country{China}}

\author{Changbin Zhao}
\email{pvthon@mail.ustc.edu.cn}
\affiliation{%
  \institution{University of Science and Technology of China}
  \city{Hefei}
  \country{China}}

\author{Jin Zhang}
\email{mjzj35723@gmail.com}
\affiliation{%
  \institution{Lanzhou University}
  \city{Lanzhou}
  \country{China}}

\author{Dexu Yu}
\email{yu.dex@northeastern.edu}
\affiliation{%
  \institution{Fenz.AI}
  \city{Palo Alto}
  \state{CA}
  \country{United States}}

\author{Tianmin Sun}
\email{tianmisun2-c@my.cityu.edu.hk}
\affiliation{%
  \institution{City University of Hong Kong}
  \city{Hong Kong}
  \country{China}}

\author{Jining Luan}
\email{202521081036@std.uestc.edu.cn}
\affiliation{%
  \institution{University of Electronic Science and Technology of China}
  \city{Chengdu}
  \country{China}}

\author{Chunxiao Li}
\email{chunxiao.li@ustc.edu.cn}
\affiliation{%
  \institution{University of Science and Technology of China}
  \city{Hefei}
  \country{China}}

\author{Hanwen Du}
\authornote{Hanwen Du and Youhua Li are the corresponding authors.}
\email{du.1128@osu.edu}
\affiliation{%
  \institution{The Ohio State University}
  \city{Columbus}
  \state{Ohio}
  \country{United States}}

\author{Youhua Li}
\authornotemark[2]
\email{youhuali2-c@my.cityu.edu.hk}
\affiliation{%
  \institution{City University of Hong Kong}
  \city{Hong Kong}
  \country{China}}

\renewcommand{\shortauthors}{Yanyan Wu et al.}

\begin{abstract}
Offline sepsis treatment aims to learn treatment policies from retrospective ICU trajectories. 
Decision Transformer-based methods provide a natural sequence modeling framework for this task, but they mainly rely on raw physiological states and historical clinician actions. 
Such raw trajectory tokens may not explicitly capture clinical semantics that clinicians use in practice, such as patient severity, recent treatment response, and current risk priorities. 
To address this limitation, we propose \textbf{ClinSDT}, an LLM-assisted Decision Transformer framework for offline sepsis treatment. 
ClinSDT uses a frozen LLM as a semantic bridge rather than a treatment decision maker. 
It constructs three clinical semantic views from ICU trajectories, namely \textsc{Profile}, \textsc{Context}, and \textsc{Advisory}, and inserts their encoded representations as independent tokens into a Decision Transformer policy. 
To support policy refinement and rollout, ClinSDT further trains a Treatment-Response Forecaster to estimate the next physiological state, SAPS-II change, and predictive uncertainty from state--action inputs. 
In the second training stage, the forecaster screens local counterfactual treatment candidates, and only low-uncertainty candidates with lower predicted SAPS-II increase are distilled into the policy after semantic regeneration. 
During inference, ClinSDT performs closed-loop treatment rollout by refreshing semantic inputs from the currently available patient state. 
Experiments on MIMIC-III and MIMIC-IV show that ClinSDT achieves consistent relative improvements over representative imitation learning, offline reinforcement learning, and Decision Transformer baselines under the same rollout protocol. Code and data are available at  \textcolor{blue}{\href{https://github.com/KAG778/ClinSDT}{ClinSDT}}.

\end{abstract}



\maketitle

\section{Introduction}

\begin{figure}[t]
    \centering
    \Description{Motivation diagram of ClinSDT showing that raw ICU trajectories contain implicit clinical semantics, which are encoded by a frozen LLM as structured context for a Decision Transformer policy.}
    \includegraphics[width=\columnwidth]{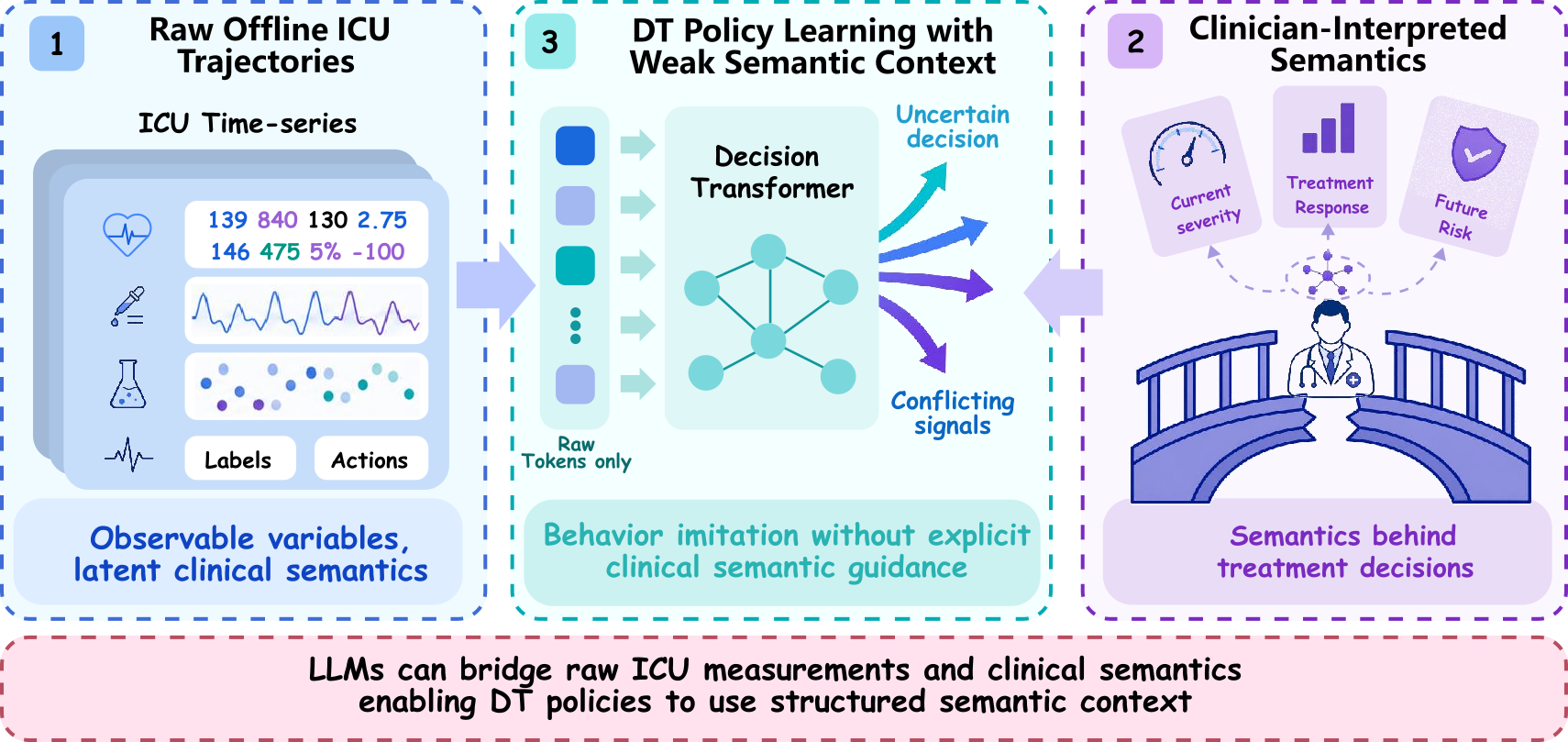}
    \caption{
    Motivation of ClinSDT.
    Raw ICU trajectories leave clinical decision semantics implicit in numerical states and actions.
    ClinSDT uses a frozen LLM to encode structured clinical context, while a Decision Transformer remains responsible for treatment action generation.
    }
    \label{fig:motivation}
\end{figure}

Sepsis treatment is a critical sequential decision-making problem in intensive care medicine~\cite{hotchkiss2003pathophysiology}. 
Patient conditions evolve over time under disease progression and treatment interventions, requiring clinicians to adjust fluid administration, vasopressor use, and other treatments based on vital signs, laboratory tests, organ function, and prior treatment history. 
Because online trial-and-error in real ICUs is costly and risky, retrospective ICU trajectories provide an important basis for learning clinical decision support policies~\cite{komorowski2018artificial,raghu2017continuous,gottesman2019guidelines}. 
Existing approaches, including behavior cloning, offline reinforcement learning, and Decision Transformer-based methods, have been used to learn treatment policies from historical trajectories~\cite{pomerleau1988alvinn,rajeswaran2017learning,levine2020offline,kumar2020conservative,chen2021decision}. 
In particular, Decision Transformer formulates offline decision-making as conditional sequence modeling and provides a natural framework for trajectory-based clinical policy learning~\cite{chen2021decision,janner2021offline}.

Despite their usefulness, most existing methods mainly learn from raw physiological states and historical clinician actions~\cite{kong2008clinical,killian2020representation,tang2021model,wang2025improving}. 
As illustrated in Figure~\ref{fig:motivation}, raw ICU measurements contain observable variables, but the clinical semantics behind treatment decisions are often implicit. 
Clinicians usually interpret these measurements in terms of patient severity, recent treatment response, and future risk priorities. 
A policy trained only on raw trajectory tokens may capture temporal patterns and behavior regularities, but it may still operate under weak semantic context. 
This limitation is especially relevant in sepsis treatment, where similar numerical states may imply different treatment priorities depending on recent responses and risk conditions.

Large language models offer a possible way to bridge raw ICU measurements and clinical semantics~\cite{singhal2023large,thirunavukarasu2023large,moor2023foundation}. 
Based on this motivation, we propose \textbf{Clinical Semantic Decision Transformer} (\textbf{ClinSDT}), an LLM-assisted Decision Transformer framework for offline sepsis treatment. 
ClinSDT uses a frozen LLM as a semantic bridge rather than a treatment decision maker. 
It constructs three clinical semantic views from ICU trajectories: \textsc{Profile} for overall patient condition, \textsc{Context} for recent treatment-response transitions, and \textsc{Advisory} for current risk priorities without recommending specific treatments. 
The encoded semantic views are inserted as independent tokens into a Decision Transformer, enabling the policy to use structured clinical context together with return-to-go, physiological states, and historical actions.

Specifically, the framework is organized around two offline training stages and one inference stage. 
First, ClinSDT learns a semantic-token policy from documented clinician actions and trains a forecaster to estimate short-term patient responses from state--action inputs. 
Second, the forecaster screens local counterfactual treatment candidates, and only low-uncertainty candidates with lower predicted SAPS-II increase are distilled into the policy after semantic regeneration. 
During inference, ClinSDT performs closed-loop treatment rollout by refreshing semantic inputs from the currently available patient state, generating an action with the policy, and forecasting the next state for subsequent decisions. Experiments on MIMIC-III and MIMIC-IV show that ClinSDT achieves consistent relative gains over representative imitation learning, offline reinforcement learning, and Decision Transformer baselines under the same rollout protocol.

We summarize our contributions as follows:
\begin{itemize}[leftmargin=*, topsep=0pt]
    \item We propose a semantic-guidance perspective for LLM-assisted offline clinical decision-making, where the LLM bridges raw ICU trajectories and clinical semantics while a constrained Decision Transformer remains responsible for treatment action generation.

    \item We construct \textsc{Profile}, \textsc{Context}, and \textsc{Advisory} semantic views and inject them as independent tokens into the Decision Transformer, providing structured clinical context beyond raw physiological states and historical actions.

    \item We introduce two-stage training that combines Semantic-Token DT learning and Treatment-Response Forecasting, then refines the policy using reliable counterfactual candidates predicted to improve outcomes.

    \item We evaluate ClinSDT on MIMIC-III and MIMIC-IV, showing consistent relative improvements over representative baselines under a closed-loop rollout protocol.
\end{itemize}

\section{Related Work}

\subsection{Offline Clinical Treatment Decision Making}

Offline clinical treatment decision making has evolved from behavior cloning
and offline reinforcement learning to conditional sequence modeling over
longitudinal trajectories~\cite{pomerleau1988alvinn,rajeswaran2017learning,
levine2020offline,fujimoto2019off,kumar2020conservative,
kostrikov2021offline,peng2019advantage}.
Killian et al.~\cite{killian2020representation} showed that patient-state
representations substantially affect policy performance in healthcare.
Decision Transformer formulates offline decision making as return-conditioned
sequence generation~\cite{chen2021decision,janner2021offline}, while
MeDT~\cite{medt2024}
adapts it to sepsis treatment through outcome and target-acuity conditioning.
Related studies further emphasize behavior support, uncertainty, model
selection, and generalization in offline policy learning~\cite{thomas2015high,
char2023correlated,tang2021model,wang2025improving}.
Despite this progress, existing methods remain primarily driven by numerical
states, actions, and target variables, leaving clinical decision
semantics---including the patient's overall condition, recent treatment
response, and current risk priorities---largely implicit.

\subsection{LLM-Empowered Clinical Policy Learning}

Biomedical language models provide reusable medical semantics
~\cite{alsentzer2019publicly,lee2020biobert}, while broader studies document
both their clinical capabilities and deployment constraints
~\cite{singhal2023large,singhal2025toward,thirunavukarasu2023large,
moor2023foundation,loni2025review}. Recent work uses LLMs to enhance clinical
representations: CARER~\cite{carer2024} aligns clinical rationales with patient
states, LEADER~\cite{leader2024} distills medical knowledge for medication
recommendation, and RePrompT~\cite{reprompt2026} integrates structured EHRs
while preserving temporal structure. In clinical offline reinforcement
learning, MORE-CLEAR~\cite{moreclear2025} fuses clinical notes with structured
measurements, whereas medR~\cite{medr2026} uses an LLM to design rewards. Beyond
healthcare, GIFT~\cite{wu2026giftllmguidedstaterewardinterface} constructs state--reward interfaces while
retaining a numerical policy, Zhu et al.~\cite{zhu2026role} integrate language
representations with numerical trajectories at the token level, and Li et
al.~\cite{li2026latent} use learned rewards to steer LLM latent states at
inference time. These studies place language-based guidance at the interface,
policy, or latent-state level. Yet role-separated semantics derived from
structured clinical trajectories have not been sufficiently explored as direct
conditions for sequential treatment policies. ClinSDT addresses this gap by
encoding \textsc{Profile}, \textsc{Context}, and \textsc{Advisory} as independent
policy tokens while retaining a Decision Transformer as the treatment decision
model.

\section{Method}

\begin{figure*}[t]
    \centering
    \Description{Overview of ClinSDT, showing how Profile, Context, and Advisory semantics are encoded by a frozen LLM, injected as independent tokens into a Decision Transformer policy, and used with treatment-response forecasting and uncertainty-constrained conservative counterfactual distillation for model-based treatment rollout.}
    \includegraphics[width=\textwidth]{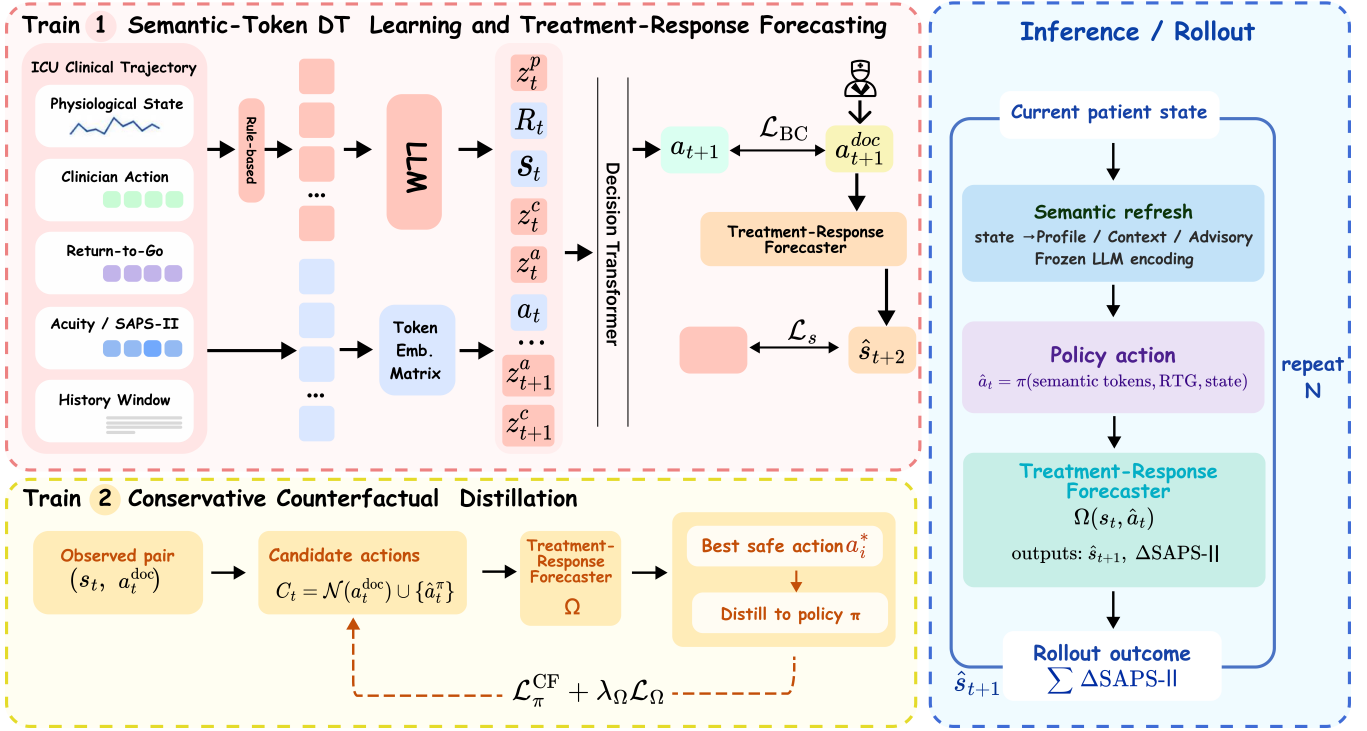}
    \caption{
    Overview of ClinSDT.
    ClinSDT encodes \textsc{Profile}, \textsc{Context}, and \textsc{Advisory} semantics from ICU trajectories with a frozen LLM and injects them as independent tokens into a Decision Transformer policy.
    Training consists of Stage I semantic-token policy and response-forecaster learning, followed by Stage II uncertainty-constrained counterfactual semantic refinement.
    During closed-loop rollout, the policy refreshes semantic inputs from each predicted patient state while retaining the retrospectively initialized target-return condition.
    The LLM provides semantic representations only; treatment actions are generated by the policy model.
    In the schematic, $\Omega\equiv F_\phi$ and $\mathcal L_s\equiv\mathcal L_{\mathrm{fr}}^{\mathrm{fact}}$.
    }
    \label{fig:framework}
\end{figure*}
ClinSDT is organized around a single design principle: the numerical trajectory
remains the basis of treatment control, while a frozen LLM supplies a structured
semantic abstraction of the clinical information available at each decision
step. As shown in Figure~\ref{fig:framework}, three role-separated semantic views
augment return-to-go, physiological states, and historical actions as
independent policy tokens. Stage I learns this semantic-token policy from
documented actions and fits a Treatment-Response Forecaster on factual
transitions. Stage II uses the forecaster to identify conservative local
counterfactual associations, regenerate their clinical semantics, and refine
the policy under factual anchoring. At inference, the policy and forecaster form
a closed loop in which each predicted transition supplies the semantic context
for the next decision. This preserves the division of labor introduced in
Section~1: the LLM represents clinical context, whereas the Decision Transformer
generates treatment actions.

\subsection{Problem Setup}

We formulate offline sepsis treatment as a finite-horizon sequential decision-making problem.
Each ICU trajectory is represented as
\begin{equation}
\tau=\{(s_t,a_t,r_t,s_{t+1})\}_{t=0}^{T-1},
\label{eq:trajectory}
\end{equation}
where \(s_t\in\mathbb{R}^{d_s}\) and \(s_{t+1}\) are consecutive physiological states, \(a_t\in\mathcal{A}\) is a discrete treatment action, and \(r_t\) is the clinical reward. 
The return-to-go is computed as
\begin{equation}
R_t=\sum_{t'=t}^{T-1}\gamma^{\,t'-t} r_{t'}, 
\qquad \gamma=0.99,
\qquad \widetilde R_t=R_t/10 .
\label{eq:rtg}
\end{equation}
We instantiate the reward as the negative one-step SAPS-II change,
\begin{equation}
r_t = -\,\Delta^{\mathrm{saps}}_t = -\big(\mathrm{SAPS\text{-}II}_{t+1}-\mathrm{SAPS\text{-}II}_t\big),
\label{eq:reward}
\end{equation}
so that a decrease in SAPS-II yields positive reward and \(R_t\) measures the
discounted cumulative SAPS-II improvement. The Transformer receives the scaled
condition \(\widetilde R_t\), while the unscaled \(R_t\) is available to the
factual \textsc{Profile} constructor. Following the retrospective DT protocol,
DT-style policies initialize this target condition from the held-out trajectory
and keep it fixed during rollout; non-DT policies do not use RTG. The learning
objective is a policy $\pi_\theta(a_t\mid H_t)$ whose history $H_t$ contains only
information available before selecting $a_t$. During factual training, $H_t$ is
formed from the observed trajectory prefix; during rollout, it is formed from
the current predicted state and previously generated transitions.

\subsection{Role-Separated Clinical Semantic State Abstraction}
\label{sec:semantic-construction}

The central representation in ClinSDT is a role-separated semantic abstraction
of the current decision state. Rather than converting the entire trajectory into
one text summary, we construct three complementary views: \textsc{Profile}
describes overall severity, organ burden, and ventilation status;
\textsc{Context} summarizes the incoming treatment-response transition; and
\textsc{Advisory} identifies current risk priorities without prescribing an
action. Their separation makes global condition, recent response, and immediate
risk independently addressable by the policy.

Three rule-based constructors, $g_p$, $g_c$, and $g_a$, instantiate these views.
Let $m_t$ denote the recorded acuity information, comprising the SAPS-II total
and organ sub-scores. For factual policy learning, the latest completed
transition before predicting $a_t$ is $(s_{t-1},a_{t-1},s_t)$.
For \(t>0\), we construct
\begin{align}
x^p_t &= g_p(s_t,m_t,R_t), \label{eq:profile_text}\\
x^c_t &= g_c(s_{t-1},a_{t-1},s_t), \label{eq:context_text}\\
x^a_t &= g_a(s_t,m_t). \label{eq:advisory_text}
\end{align}
The construction follows decision-time availability. Equations~\eqref{eq:profile_text}--\eqref{eq:advisory_text}
use the current factual state and the incoming transition, so the action being
predicted never appears in its own semantic input. At $t=0$, $g_c$ uses an
initial-state template because no incoming transition exists. At rollout states,
recorded acuity is unavailable: $g_p$ and $g_a$ use state-only variants, the
\textsc{Profile} text omits the return target, and $g_c$ summarizes the generated
incoming transition; the separate fixed RTG token is unchanged. Stage II uses a
different timing convention: hindsight \textsc{Context} and \textsc{Advisory}
come from the predicted counterfactual next state, while \textsc{Profile} remains
anchored to the factual pre-decision condition (Section~\ref{sec:ccd}). Each constructor maps thresholded
physiological features to a short qualitative description and samples a surface
realization within the corresponding clinical category; consequently, the
views expose qualitative structure rather than raw numerical values.
Appendix~\ref{app:prompts} provides the input regimes and examples.

A frozen LLM encoder maps the constructed descriptions into semantic embeddings:
\begin{align}
z^p_t &= E_{\mathrm{LLM}}(x^p_t), 
\label{eq:profile_embed}\\
z^c_t &= E_{\mathrm{LLM}}(x^c_t), 
\label{eq:context_embed}\\
z^a_t &= E_{\mathrm{LLM}}(x^a_t).
\label{eq:advisory_embed}
\end{align}
We use frozen Qwen2.5-0.5B-Instruct for $E_{\mathrm{LLM}}$ and mean-pool its
last-layer hidden states into an $896$-dimensional embedding for each view. The
three embeddings are passed to the policy as semantic state variables; the LLM
neither produces action logits nor recommends treatments. The constructors and
frozen encoder together form the semantic-guidance pipeline evaluated in this
work.

\subsection{Stage I: Semantic-Token Policy and Response Forecasting}

Stage I establishes the two components used throughout ClinSDT: a
Semantic-Token Decision Transformer learned from documented treatment behavior,
and a Treatment-Response Forecaster learned from factual state--action
transitions. The former maps the semantic abstraction and numerical trajectory
to actions; the latter provides the response model required by Stage II and
closed-loop rollout.

\paragraph{\bfseries\itshape Semantic-token policy learning.}
ClinSDT injects the three clinical semantic views into the Decision Transformer as independent tokens. 
For each time step \(t\), we distinguish the complete decision block from the
pre-action prefix used to predict the current action:
\begin{equation}
\begin{aligned}
b_t&=[z^p_t,\widetilde R_t,z^c_t,z^a_t,s_t,a_t],\\
b_t^-&=[z^p_t,\widetilde R_t,z^c_t,z^a_t,s_t].
\end{aligned}
\label{eq:decision_block}
\end{equation}
where \(z^p_t\), \(z^c_t\), and \(z^a_t\) are the \textsc{Profile},
\textsc{Context}, and \textsc{Advisory} embeddings. Their separate positions
preserve the global-condition, recent-transition, and risk-priority roles.

Given a context window of length \(L\), define the sequence of complete
historical blocks available before predicting \(a_t\) as
\begin{equation}
\mathcal H_t^{(L)}
:=
\big[b_j\big]_{j=\max(0,t-L+1)}^{t-1},
\qquad
B_t^-=[\mathcal H_t^{(L)},b_t^-],
\label{eq:context_window}
\end{equation}
where $\mathcal H_0^{(L)}$ is empty. Padding and attention masking handle the
beginning of a trajectory. Complete historical blocks contain documented
actions, whereas the current block terminates at $s_t$; hence the causal
sequence exposes past actions but not $a_t$ itself. All tokens are projected
into a shared hidden space and processed by a causal Transformer.
The policy head reads the hidden state at the current state token and outputs action logits:
\begin{equation}
\ell_t=f_{\theta}(h^s_t)\in\mathbb{R}^{|\mathcal{A}|}.
\label{eq:policy_logits}
\end{equation}
The action probability is given by
\begin{equation}
p_{\theta}(a\mid B_t^-)=\mathrm{softmax}(\ell_t)[a].
\label{eq:action_prob}
\end{equation}
The policy is trained by behavior cloning on documented clinician actions:
\begin{equation}
\mathcal{L}_{\mathrm{BC}}
=
-\log p_{\theta}(a_t^{\mathrm{doc}}\mid B_t^-).
\label{eq:bc_loss}
\end{equation}

\paragraph{\bfseries\itshape Response forecasting.}
The dropout-MLP forecaster \(F_{\phi}\) predicts a diagonal-Gaussian next state
and the SAPS-II change from factual transitions:
\begin{equation}
F_{\phi}(s_t,a_t)=\big(\mu_{\phi}(s_t,a_t),\ \log\sigma_{\phi}(s_t,a_t),\ \hat{d}^{\mathrm{saps}}_{\phi}(s_t,a_t)\big),
\label{eq:forecaster_output}
\end{equation}
where \(\sigma^2_{\phi}=\exp(2\log\sigma_{\phi})\). The forecaster uses only
numerical state--action inputs and is trained with
\begin{equation}
\mathcal{L}^{\mathrm{fact}}_{\mathrm{fr}}
=
\mathrm{NLL}\!\big(s_{t+1}\mid \mu_{\phi},\sigma^2_{\phi}\big)\big|_{a_t^{\mathrm{doc}}}
+
\big(\hat{d}^{\mathrm{saps}}_{\phi}(s_t,a_t^{\mathrm{doc}})-\Delta^{\mathrm{saps}}_t\big)^2 ,
\label{eq:forecaster_fact_loss}
\end{equation}
For candidate screening, \(M=10\) MC-dropout passes provide the averaged next
state
\begin{equation}
\hat{s}_{t+1}
=
\bar{F}^{s}_{\phi}(s_t,a_t)
=
\frac{1}{M}\sum_{m=1}^{M}\mu_{\phi}^{(m)} .
\label{eq:mc_state_mean}
\end{equation}
The same passes give
\(\bar d_{\phi}^{\mathrm{saps}}(s_t,a_t)=
\operatorname{avg}_{m}\hat d_{\phi}^{\mathrm{saps},(m)}(s_t,a_t)\)
and the scalar predictive-dispersion score
\begin{equation}
\hat{\sigma}_{\phi}(s_t,a_t)
=
\mathrm{mean}_{j}\!\left(
\underbrace{\mathrm{std}_{m}\!\left[\mu_{\phi,j}^{(m)}\right]}_{\text{epistemic}}
+
\underbrace{\frac{1}{M}\sum_{m=1}^{M}\sigma_{\phi,j}^{(m)}}_{\text{aleatoric}}
\right),
\label{eq:mc_uncertainty}
\end{equation}
where \(j\) indexes state dimensions. Candidates with
\(\hat{\sigma}_{\phi}(s_t,a_t)\geq\sigma_{\mathrm{thr}}\) are removed;
\(F^{s}_{\phi}\) denotes one stochastic dropout forward pass, while
\(\bar F^{s}_{\phi}\) denotes the MC average.

\subsection{Stage II: Conservative Counterfactual Semantic Refinement}
\label{sec:ccd}

Stage II uses the factual response model to refine the semantic-token policy
without performing unconstrained action search. For each factual decision state,
it evaluates a behavior-centered treatment neighborhood together with the
current policy proposal, retains only low-uncertainty candidates, and distills an
accepted predicted-improving association after regenerating its clinical
semantics.

\paragraph{\bfseries\itshape Candidate construction and screening.}
For each factual state, we construct a candidate set using local neighbors of the documented clinician action and the current policy action:
\begin{equation}
\mathcal{C}_t
=
\mathcal{N}_{\rho}(a_t^{\mathrm{doc}})
\cup
\{\hat{a}^{\pi}_t\},
\qquad
\hat{a}^{\pi}_t=\arg\max_a \ell_t[a],
\label{eq:candidate_set}
\end{equation}
Let \(u(a)=\lfloor a/5\rfloor\) and \(v(a)=a\bmod 5\) denote the fluid and
vasopressor coordinates of action \(a\), respectively. The implemented
two-dimensional grid neighborhood is
\begin{equation}
\mathcal{N}_{\rho}(a^{\mathrm{doc}})
=\left\{5i+j:\begin{array}{l}
0\le i,j<5,\\[-2pt]
|i-u(a^{\mathrm{doc}})|\le\rho,\\[-2pt]
|j-v(a^{\mathrm{doc}})|\le\rho
\end{array}\right\},\qquad \rho=2,
\label{eq:neighborhood}
\end{equation}
on the \(5\times5\) fluid--vasopressor grid; the policy proposal need not be
grid-local. Let \(\mathcal C_t^{\mathrm{valid}}=\{a\in\mathcal C_t:
\hat\sigma_{\phi}(s_t,a)<\sigma_{\mathrm{thr}}\}\). If this set is nonempty,
we select \(\tilde a_t=\arg\min_{a\in\mathcal C_t^{\mathrm{valid}}}
\bar d_{\phi}^{\mathrm{saps}}(s_t,a)\); otherwise, \(\tilde a_t=\varnothing\).
With \(\sigma_{\mathrm{thr}}=2.0\), the candidate is accepted only if its
predicted one-step SAPS-II change is lower than the documented action; the
initial step is excluded because it has no observed incoming transition:
\begin{equation}
a_t^*
=
\begin{cases}
\tilde{a}_t,
&
\tilde a_t\neq\varnothing
\ \wedge\ 
\bar d^{\mathrm{saps}}_{\phi}(s_t,a_t^{\mathrm{doc}})
>
\bar d^{\mathrm{saps}}_{\phi}(s_t,\tilde{a}_t)
\ \wedge\ t>0,\\
\varnothing, & \text{otherwise}.
\end{cases}
\label{eq:accepted_action}
\end{equation}
Together, locality, uncertainty filtering, and comparison with the documented
action make this a conservative one-step refinement rule.

\paragraph{\bfseries\itshape Counterfactual semantic target construction.}
An accepted association $s_t\xrightarrow{a_t^*}\hat s_{t+1}^{\mathrm{cf}}$
is converted into a training-only semantic target. The forecaster first predicts
the counterfactual next state:
\begin{equation}
\hat{s}^{\mathrm{cf}}_{t+1}
=
\bar F_{\phi}^{s}(s_t,a_t^*).
\label{eq:cf_next_state}
\end{equation}
The predicted transition then determines the response-oriented semantic views:
\begin{align}
z^{c,\mathrm{cf}}_{t+1}
&=
E_{\mathrm{LLM}}
\left(
g_c(s_t,a_t^*,\hat{s}^{\mathrm{cf}}_{t+1})
\right),
\label{eq:cf_context}\\
z^{a,\mathrm{cf}}_{t+1}
&=
E_{\mathrm{LLM}}
\left(
g_a(\hat{s}^{\mathrm{cf}}_{t+1})
\right).
\label{eq:cf_advisory}
\end{align}
The \textsc{Profile} token remains tied to the factual pre-decision condition:
\begin{equation}
z^p_t
=
E_{\mathrm{LLM}}(g_p(s_t,m_t,R_t)).
\label{eq:cf_profile}
\end{equation}
Together, the pre-decision \textsc{Profile} and the predicted response views
form the counterfactual distillation input
\begin{equation}
B_{t\rightarrow t+1}^{\mathrm{cf}}
=
[z^p_t,\widetilde R_t,z^{c,\mathrm{cf}}_{t+1},z^{a,\mathrm{cf}}_{t+1},s_t].
\label{eq:cf_block}
\end{equation}
The full block appends a causally masked action token after the state token. This
construction treats the predicted consequence as supervision for the accepted
association, not as an online input for choosing the same action. Chronological
rollout restores the ordinary incoming-transition semantics: after $\hat a_k$
is selected and $(\hat s_k,\hat a_k,\hat s_{k+1})$ is generated, that transition
constructs \textsc{Context} for $\hat a_{k+1}$, while \textsc{Profile} and
\textsc{Advisory} are refreshed from $\hat s_{k+1}$. Appendix~\ref{app:training_eval}
summarizes this timing across stages.

\paragraph{\bfseries\itshape Policy distillation.}
The regenerated semantic target updates the policy toward the accepted action:
\begin{equation}
\ell_{t\rightarrow t+1}^{\mathrm{cf}}
=
f_\theta(B_{t\rightarrow t+1}^{\mathrm{cf}})
\in\mathbb{R}^{|\mathcal{A}|},
\label{eq:cf_logits}
\end{equation}
\begin{equation}
\mathcal{L}_{\mathrm{cf}}
=
\mathrm{MSE}
\left(
\ell_{t\rightarrow t+1}^{\mathrm{cf}},
\mathbf{e}(a_t^*)
\right),
\label{eq:cf_loss}
\end{equation}
where \(\mathbf{e}(a_t^*)\) is one-hot; the implemented MSE loss is applied only
to accepted candidates, with \(\mathcal L_{\mathrm{cf}}=0\) otherwise.

\paragraph{\bfseries\itshape Forecaster anchoring and bootstrap regularization.}
During Stage II, the forecaster continues to receive factual supervision through \(\mathcal{L}^{\mathrm{fact}}_{\mathrm{fr}}\). 
For an accepted candidate, we further add a confidence-gated bootstrap regularization term:
\begin{equation}
\mathcal{L}_{\mathrm{boot}}
=
\left\|
F_{\phi}^{s}(s_t,a_t^*)
-
\mathrm{sg}
\left[
\bar{F}_{\phi}^{s}(s_t,a_t^*)
\right]
\right\|_2^2,
\label{eq:boot_loss}
\end{equation}
where \(\bar{F}_{\phi}^{s}\) is the MC-dropout average and
\(\mathrm{sg}[\cdot]\) stops gradients. The confidence score and bootstrap
weight are
\begin{equation}
c_t^{\mathrm{conf}}
=
\mathrm{clip}
\left(
1-\frac{\hat{\sigma}_{\phi}(s_t,a_t^*)}{\sigma_{\mathrm{thr}}},
0,1
\right),
\label{eq:confidence}
\end{equation}
\begin{equation}
\alpha_t
=
\mathbb{I}[c_t^{\mathrm{conf}}\ge\kappa]\cdot c_t^{\mathrm{conf}}\cdot
\max
\left(
\alpha_{\min},
1-\frac{n}{N}
\right),
\label{eq:boot_weight}
\end{equation}
where \(\kappa=0.5\), \(\alpha_{\min}=0.3\), \(N=5\times10^4\), and \(n\) is the global training step. These quantities are defined for accepted candidates; when none is accepted, we set \(\mathcal L_{\mathrm{boot}}=\alpha_t=0\).

The Stage II forecaster objective combines factual anchoring and the gated
bootstrap term,
\(\mathcal L_{\Omega}:=\mathcal L^{\mathrm{fact}}_{\mathrm{fr}}+
\alpha_t\mathcal L_{\mathrm{boot}}\). The Stage II objective is
\begin{equation}
\mathcal{L}_{\mathrm{stage2}}
=
\mathcal{L}_{\mathrm{cf}}
+
\lambda_{\Omega}\mathcal L_{\Omega},
\qquad \lambda_{\Omega}=0.5.
\label{eq:stage2_loss}
\end{equation}
Stage II retains no behavior-cloning or KL term. Its two forms of conservatism
act on different components: the local neighborhood, uncertainty gate, and
documented-action comparison constrain the policy target, while factual
anchoring and confidence-gated self-consistency stabilize the forecaster.

\subsection{Inference: Closed-Loop Semantic-Guided Rollout}

Inference applies the learned components in chronological order. At each rollout
step, the currently available state and incoming transition are abstracted into
\textsc{Profile}, \textsc{Context}, and \textsc{Advisory}, encoded by the frozen
LLM, and passed to the policy together with the numerical decision prefix. With
the current action token absent, the policy selects
\begin{equation}
\hat{a}_k=\arg\max_{a\in\mathcal{A}}\ell_k[a],
\label{eq:inference_action}
\end{equation}
from the input
\begin{equation}
[z^p_k,\widetilde R^{\mathrm{obs}}_0,z^c_k,z^a_k,\hat s_k],
\label{eq:inference_input}
\end{equation}
where \(\widetilde R^{\mathrm{obs}}_0\) is initialized from the held-out trajectory
and remains fixed for every rollout step \(k\).

At step 0, the semantic abstraction uses the recorded initial acuity information.
Subsequent steps use state-only \textsc{Profile} and \textsc{Advisory} templates
and a \textsc{Context} summary of the generated incoming transition. Given
$\hat{s}_k$ and $\hat{a}_k$, the forecaster advances the rollout by
\begin{equation}
\hat{s}_{k+1}
=
\bar{F}_{\phi}^{s}(\hat{s}_k,\hat{a}_k).
\label{eq:rollout_state_update}
\end{equation}
Here $\bar{F}_{\phi}^{s}$ is the MC-dropout averaged state prediction. The
resulting transition refreshes the semantic abstraction before the next action,
while the held-out-trajectory RTG initialized at step 0 remains fixed.
For method \(i\), the reported estimate is
\(\mathrm{Est}^{(i)}_H=-\sum_{k=0}^{H-1}
\bar d^{\mathrm{saps}}_{\phi_i}(\hat s_k,\hat a_k)\). Higher values correspond
to a lower cumulative SAPS-II change predicted by the corresponding forecaster.
Repeating this cycle for horizon $H$ produces the closed-loop estimate used in
Section~4. Thus the semantic abstraction evolves with the generated patient
trajectory, while action generation remains entirely under the learned policy.

\section{Experiments}

We evaluate ClinSDT from four perspectives: overall comparison with
representative baselines, the contribution of the three clinical semantic views
and their token organization, the effect of Stage II refinement and
semantic-token diagnostics, and rollout consistency with behavior-support
diagnostics.

\begin{description}[leftmargin=0.15cm, labelwidth=0.5cm, labelsep=0.2cm, itemsep=2pt, topsep=3pt]
    \item[\textbf{RQ1}:] How does ClinSDT compare with representative baselines under the same rollout procedure?
    \item[\textbf{RQ2}:] What do the three semantic views and their independent-token organization contribute?
    \item[\textbf{RQ3}:] How does Stage II affect the rollout estimates, and what do the semantic-token diagnostics reveal?
    \item[\textbf{RQ4}:] Are the rollout trends consistent across horizons and compatible with the logged action distribution?
\end{description}

\begin{table*}[t]
\centering
\scriptsize
\setlength{\tabcolsep}{4.2pt}
\renewcommand{\arraystretch}{0.88}
\caption{
Overall model-based rollout results on MIMIC-III and MIMIC-IV.
We compare ClinSDT with representative baselines from imitation learning, offline RL, value-based RL, and Decision Transformer-style policy learning.
The metric is negative cumulative forecaster-predicted SAPS-II change. All
methods follow the same rollout procedure, while each uses its own learned
forecaster; the values are therefore protocol-relative point estimates rather
than scores on a shared independent scale. \textbf{Bold} and
\underline{underlining} indicate only the numerical ordering within the table.
}
\label{tab:main_results_mimic3_mimic4_rollout}
\resizebox{0.8\textwidth}{!}{%
\begin{tabular}{llcccccccc}
\toprule
\multirow{2}{*}{\textbf{Dataset}} 
& \multirow{2}{*}{\textbf{Model}}
& \multicolumn{4}{c}{\textbf{5-step rollout}} 
& \multicolumn{4}{c}{\textbf{10-step rollout}} \\
\cmidrule(lr){3-6} \cmidrule(lr){7-10}
& 
& \textbf{HIGH} 
& \textbf{MID} 
& \textbf{LOW} 
& \textbf{ALL}
& \textbf{HIGH} 
& \textbf{MID} 
& \textbf{LOW} 
& \textbf{ALL} \\
\midrule

\multirow{6}{*}{\textbf{MIMIC-III}}
& BC  
& 3.24 & -1.51 & -5.59 & -0.59 
& 5.39 & -1.55 & -7.80 & -0.22 \\
& BCQ 
& 3.17 & -1.30 & \underline{-3.88} & -0.35 
& 5.99 & -1.17 & \underline{-5.59} & 0.34 \\
& DQN 
& 6.42 & \underline{-1.05} & -5.03 & \underline{0.57} 
& \underline{13.17} & \underline{0.10} & -6.74 & \underline{2.94} \\
& DT  
& \textbf{7.01} & -1.57 & -5.93 & 0.30 
& 12.96 & -1.40 & -8.98 & 1.72 \\
& IQL 
& 2.86 & -2.13 & -6.13 & -1.14 
& 4.39 & -2.58 & -9.25 & -1.28 \\
\rowcolor{gray!15}
& \textbf{Ours} 
& \underline{6.91} 
& \textbf{-0.44} 
& \textbf{-1.43} 
& \textbf{1.36} 
& \textbf{17.74} 
& \textbf{5.16} 
& \textbf{10.77} 
& \textbf{8.75} \\

\midrule

\multirow{6}{*}{\textbf{MIMIC-IV}}
& BC  
& 0.55 & -0.80 & -0.62 & -0.26 
& 0.34 & -1.61 & -1.23 & -0.79 \\
& BCQ 
& \underline{1.82} & -0.66 & -0.93 & 0.13 
& \textbf{2.59} & -1.02 & -1.53 & \underline{0.10} \\
& DQN 
& 1.38 & \underline{-0.43} & \underline{-0.50} & 0.19 
& 1.97 & \underline{-0.95} & \underline{-1.06} & 0.06 \\
& DT  
& \textbf{2.17} & -0.74 & -0.95 & \underline{0.23} 
& \underline{2.48} & -1.90 & -2.18 & -0.43 \\
& IQL 
& 0.17 & -0.84 & -0.77 & -0.46 
& -1.23 & -2.15 & -1.62 & -1.65 \\
\rowcolor{gray!15}
& \textbf{Ours} 
& 0.65 
& \textbf{0.21} 
& \textbf{0.17} 
& \textbf{0.35} 
& 0.76 
& \textbf{0.30} 
& \textbf{0.27} 
& \textbf{0.45} \\

\bottomrule
\end{tabular}%
}
\end{table*}
\paragraph{\bfseries\itshape Datasets.}
We evaluate on offline ICU sepsis treatment trajectories from MIMIC-III and MIMIC-IV~\cite{johnson2016mimic,johnson2023mimic}. 
Each trajectory contains physiological measurements, historical treatment actions, and subsequent patient-condition changes. 
Following common sepsis treatment settings, each state is represented by a 45-dimensional physiological feature vector, and actions are discretized according to intravenous fluid and vasopressor levels. 
For subgroup analysis, patients are divided into \textbf{HIGH}, \textbf{MID}, and \textbf{LOW} acuity groups according to initial SAPS-II scores. 
Because SAPS-II distributions differ across datasets, we use dataset-specific thresholds while keeping the same three-level acuity structure.

\paragraph{\bfseries\itshape Evaluation metric.}
We use forecaster-predicted SAPS-II change as the proxy quantity for model-based
rollout analysis. Let \(\bar d^{\mathrm{saps}}_{\phi_i,t}\) denote the SAPS-II
change predicted at step \(t\) by method \(i\)'s MC-dropout averaged forecaster.
For rollout horizon \(H\), the cumulative predicted change is
\begin{equation}
\Delta \widehat{\mathrm{SAPS}}_{i,H}
=
\sum_{t=0}^{H-1}
\bar d^{\mathrm{saps}}_{\phi_i}
(\hat{s}_{i,t},\hat{a}_{i,t}).
\end{equation}
We report its negative as the method-specific rollout estimate,
\begin{equation}
\mathrm{Est}_{i,H}
=
-\Delta \widehat{\mathrm{SAPS}}_{i,H},
\end{equation}
and average \(\mathrm{Est}_{i,H}\) over the corresponding test trajectories.
Higher values indicate a more favorable estimate under that method's learned
forecaster, not independently observed patient improvement. 
We report both $5$-step and $10$-step rollouts. 
All methods use the same initial trajectories and rollout horizons, but each is
evaluated with its own Stage II-updated forecaster. DT-style policies initialize
RTG from the held-out trajectory and keep it fixed; non-DT methods use no RTG.
The scores are therefore method-specific descriptive estimates, not policy
values from a shared clinical environment.

\paragraph{\bfseries\itshape Baselines.}
We compare ClinSDT with representative baselines from imitation learning, value-based RL, offline RL, and sequence-modeling-based policy learning. 
\textbf{BC}~\cite{pomerleau1988alvinn} imitates logged clinician actions. 
\textbf{DQN}~\cite{mnih2015human} is a value-based RL baseline trained under the same offline data and rollout protocol. 
\textbf{BCQ}~\cite{fujimoto2019off} constrains action selection to the support of the behavior data. 
\textbf{IQL}~\cite{kostrikov2021offline} performs implicit offline policy optimization without explicit out-of-distribution action evaluation. 
\textbf{DT}~\cite{chen2021decision} is a return-conditioned Decision Transformer trained on return, state, and action sequences. 

\subsection{Overall Model-Based Rollout Results (RQ1)}

\begin{figure}[t]
    \centering
    \includegraphics[width=\columnwidth]{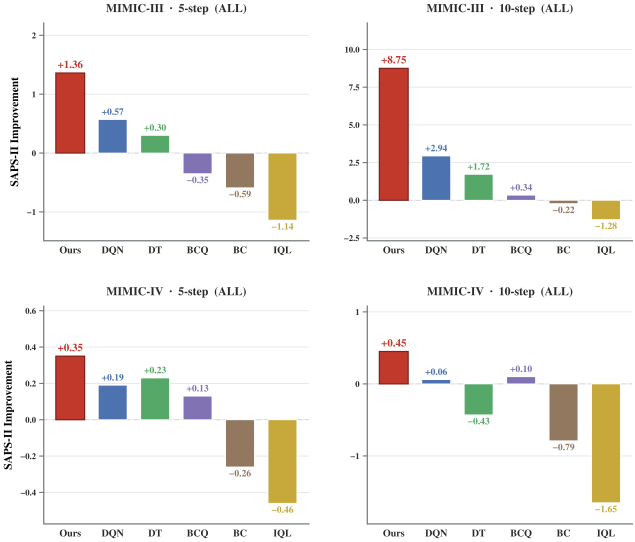}
    \caption{
    \textbf{Overall model-based rollout results on MIMIC-III and MIMIC-IV.}
    ClinSDT has the highest ALL-patient point estimate in all four reported
    dataset--horizon settings under the implemented rollout protocol.
    }
    \Description{
    A bar chart showing method-specific rollout estimates for ClinSDT and the baseline methods on MIMIC-III and MIMIC-IV under 5-step and 10-step settings. 
    The chart summarizes negative cumulative forecaster-predicted SAPS-II changes for all patients.
    }
    \label{fig:saps2_comparison}
\end{figure}

Table~\ref{tab:main_results_mimic3_mimic4_rollout} compares ClinSDT with the
baselines under $5$-step and $10$-step model rollouts. ClinSDT obtains the
highest ALL-patient point estimate in all four dataset--horizon settings. On
MIMIC-III, it reaches $+1.36$ and $+8.75$, compared with the strongest baseline
estimates of $+0.57$ and $+2.94$ at five and ten steps, respectively; at ten
steps, it also has the highest point estimate in all three acuity strata. On
MIMIC-IV, ClinSDT obtains $+0.35$ and $+0.45$ and remains highest on the
ALL-patient estimate at both horizons. Figure~\ref{fig:saps2_comparison} summarizes this
cross-dataset pattern. \textbf{ClinSDT shows a consistent relative advantage in the
ALL-patient estimates across datasets and horizons under the implemented
protocol; RQ4 defines the scope of these estimates.}

\subsection{Semantic Factors and Token Organization (RQ2)}

RQ2 explores two design choices: the content of the three clinical views and
their organization in the policy input.

\paragraph{\bfseries\itshape Semantic-view variants.}
Table~\ref{tab:rq2-view} reports removal and single-view variants on MIMIC-III.
The full model obtains the highest $10$-step ALL estimate of $+8.75$. 
Removing \textsc{Profile}, \textsc{Context}, or \textsc{Advisory} reduces the score to $+0.84$, $+1.37$, and $+1.10$, respectively. 
The reductions indicate that the complete configuration benefits from combining
the three roles. Single-view results show role specialization: \textsc{Advisory}
is strongest on the $5$-step ALL metric, whereas \textsc{Context} is strongest
in the $10$-step HIGH stratum. Thus, no single view explains the full model's
long-horizon ALL estimate.

\begin{table}[t]
\centering
\scriptsize
\setlength{\tabcolsep}{3pt}
\renewcommand{\arraystretch}{1.0}
\caption{
Semantic-view variants on MIMIC-III under 5-step and 10-step rollouts.
Results are method-specific negative cumulative forecaster-predicted SAPS-II
changes. \textbf{Bold} and \underline{underlining} indicate only the highest and
second-highest numerical point estimates.
}
\label{tab:rq2-view}
\begin{tabular}{cccccccccc}
\toprule
\multirow{2}{*}{\textbf{Group}} & \multirow{2}{*}{\textbf{Variant}}
& \multicolumn{4}{c}{\textbf{5-step rollout}} 
& \multicolumn{4}{c}{\textbf{10-step rollout}} \\
\cmidrule(lr){3-6} \cmidrule(lr){7-10}
& & \makecell{\textbf{HIGH}} & \makecell{\textbf{MID}} & \makecell{\textbf{LOW}} & \makecell{\textbf{ALL}}
  & \makecell{\textbf{HIGH}} & \makecell{\textbf{MID}} & \makecell{\textbf{LOW}} & \makecell{\textbf{ALL}} \\
\midrule
\multirow{4}{*}{Removal}
& w/o Profile  & 3.14 & -0.54 & -4.96 & 0.09 & 2.52 & 0.30 & -0.03 & 0.84 \\
& w/o Context  & 5.66 & -0.43 & -4.97 & 0.80 & 5.92 & -0.14 & -0.68 & 1.37 \\
& w/o Advisory & 5.97 & -1.30 & -7.94 & 0.09 & 4.26 & 0.05 & -0.28 & 1.10 \\
& w/o Semantics & 7.06 & -2.48 & -9.69 & -0.56 & 4.82 & 0.41 & 0.39 & 1.53 \\
\midrule
\multirow{3}{*}{Single-view}
& only Profile  & \underline{8.45} & -0.79 & -3.67 & \underline{1.36} & 1.68 & -0.94 & -0.66 & -0.25 \\
& only Context  & 2.52 & \underline{-0.13} & -6.42 & 0.10 & \textbf{24.12} & \underline{1.56} & \underline{1.75} & \underline{7.31} \\
& only Advisory & \textbf{9.19} & \textbf{0.71} & \textbf{-0.43} & \textbf{2.79} & 6.28 & 0.43 & 0.43 & 1.91 \\
\midrule
\rowcolor{gray!15}
Full & \textbf{Ours} & 6.91 & -0.44 & \underline{-1.43} & \underline{1.36} & \underline{17.74} & \textbf{5.16} & \textbf{10.77} & \textbf{8.75} \\
\bottomrule
\end{tabular}
\end{table}

\begin{figure}[t]
    \centering
    \includegraphics[width=\columnwidth]{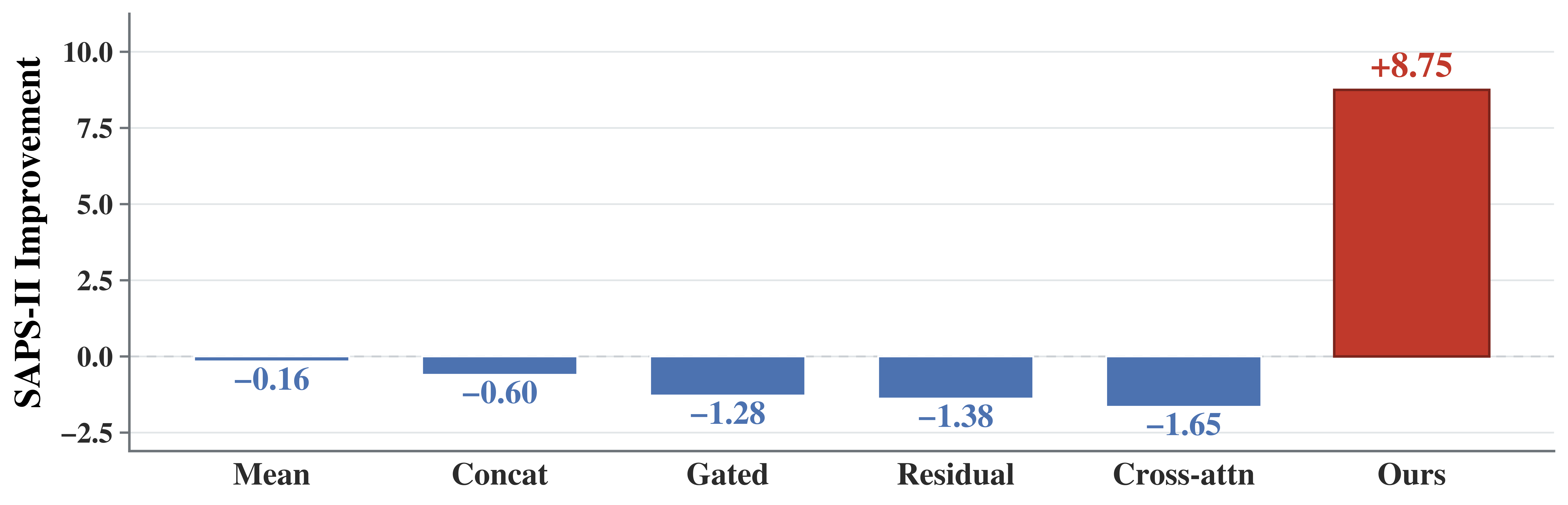}
    \caption{
    \textbf{Token-fusion variants on MIMIC-III.}
    The separated-token design has the highest $10$-step point estimate among the complete fusion configurations.
    }
    \Description{
    A bar chart comparing different semantic token-fusion designs on MIMIC-III under the 10-step rollout.
    The separated-token design has the highest numerical estimate across the reported groups under its learned forecaster.
    }
    \label{fig:token_fusion_variants}
\end{figure}

\paragraph{\bfseries\itshape Token-fusion variants.}
Table~\ref{tab:rq2-fusion} compares different ways of integrating the semantic views. 
Mean pooling, concatenation, gated fusion, residual fusion, and cross-attention
produce $10$-step ALL estimates from $-0.16$ to $-1.65$, whereas the separated-token
ClinSDT configuration reaches $+8.75$. Preserving the clinical roles as
independent tokens allows the policy to combine them contextually. As the
variants instantiate different sequence layouts, this experiment compares the
complete fusion configurations.
Figure~\ref{fig:token_fusion_variants} visualizes the same comparison.

\textbf{RQ2 supports complementary semantic views retained as
independent tokens in the complete ClinSDT configuration.}

\begin{table}[tbp]
\centering
\scriptsize
\setlength{\tabcolsep}{4pt}
\renewcommand{\arraystretch}{0.90}
\caption{
Comparison of token-fusion designs on MIMIC-III under the 10-step rollout.
Higher values indicate a more favorable estimate under the corresponding learned forecaster.
}
\label{tab:rq2-fusion}
\resizebox{0.98\columnwidth}{!}{
\begin{tabular}{llcccc}
\toprule
\textbf{Variant} & \textbf{Design}
& \textbf{HIGH} & \textbf{MID} & \textbf{LOW} & \textbf{ALL} \\
\midrule
Mean       & zero-param pooling  & 0.49  & -0.24 & -1.78  & -0.16 \\
Concat     & one projection      & 0.62  & -0.35 & -7.43  & -0.60 \\
Gated      & two projections     & 4.08  & -1.19 & -21.50 & -1.28 \\
Residual   & residual projection & -0.02 & -1.34 & -6.78  & -1.38 \\
Cross-attn & MHA + LN            & -0.94 & -1.48 & -5.89  & -1.65 \\
\midrule
\rowcolor{gray!15}
\textbf{Ours} & 6-token block (3 semantic) & \textbf{17.74} & \textbf{5.16} & \textbf{10.77} & \textbf{8.75} \\
\bottomrule
\end{tabular}
}
\end{table}

\subsection{Counterfactual Refinement and Semantic-Token Diagnostics (RQ3)}

RQ3 examines the Stage I-to-Stage II change and then uses ablations and attention
patterns as token-level diagnostics.

\begin{figure}[t]
    \centering
    \Description{Line plot showing the Stage I-to-Stage II change in method-relative rollout estimates on MIMIC-III. The y-axis is the negative cumulative SAPS-II change predicted by each method's own forecaster, so higher values are more favorable under that forecaster.}
    \includegraphics[width=\columnwidth]{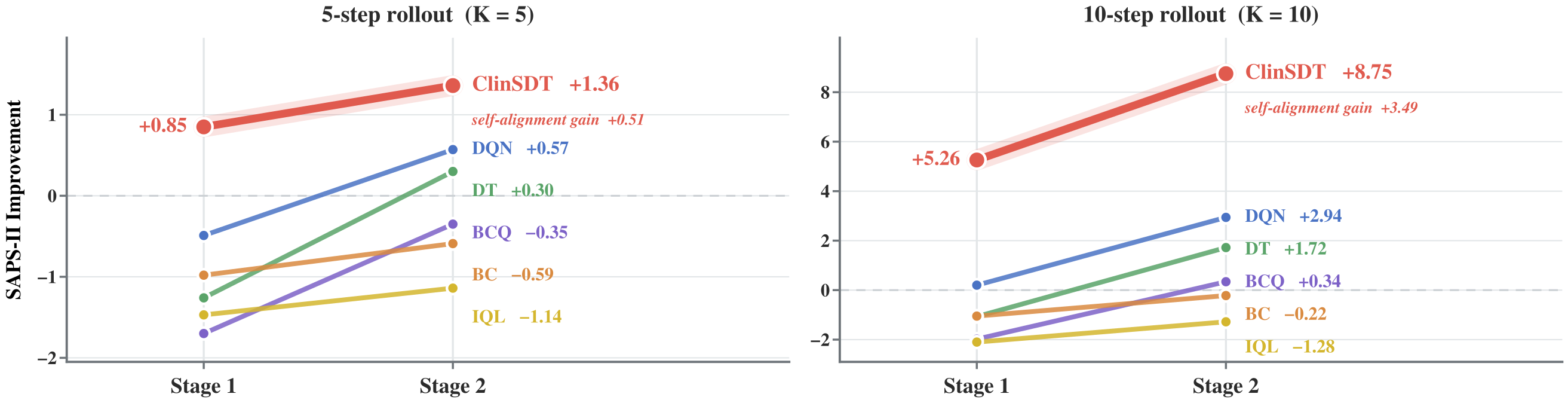}
    \caption{
    Stage I-to-Stage II change in the method-relative rollout estimate on MIMIC-III. The y-axis is defined as the negative cumulative SAPS-II change predicted by each method's own forecaster; higher is more favorable under that forecaster.
    }
    \label{fig:stage2_refinement}
\end{figure}

\paragraph{\bfseries\itshape Stage II refinement.}
Figure~\ref{fig:stage2_refinement} and
Table~\ref{tab:stage2_improvement_mimic3} report the Stage I-to-Stage II change
on MIMIC-III. ClinSDT changes by $+0.51$ on the $5$-step ALL estimate and by
$+3.49$ at ten steps, the largest reported $10$-step ALL change; DT has the
largest $5$-step ALL change. The longer-horizon result is consistent with the
intended counterfactual refinement. Since Stage II updates both components, the
values characterize the complete policy--forecaster system.

\begin{table}[t]
\centering
\scriptsize
\setlength{\tabcolsep}{2pt}
\renewcommand{\arraystretch}{0.88}
\caption{
Stage I-to-Stage II change in each method's forecaster-based estimate on MIMIC-III.
}
\label{tab:stage2_improvement_mimic3}
\resizebox{\columnwidth}{!}{
\begin{tabular}{lcccccccc}
\toprule
\multirow{2}{*}{\textbf{Model}}
& \multicolumn{4}{c}{\textbf{5-step estimate change}}
& \multicolumn{4}{c}{\textbf{10-step estimate change}} \\
\cmidrule(lr){2-5} \cmidrule(lr){6-9}
& \textbf{HIGH} & \textbf{MID} & \textbf{LOW} & \textbf{ALL}
& \textbf{HIGH} & \textbf{MID} & \textbf{LOW} & \textbf{ALL} \\
\midrule
DQN & $2.93$ & $0.45$ & $0.28$ & $1.06$ & $6.76$ & $1.41$ & $1.07$ & $2.75$ \\
DT  & $3.79$ & $0.76$ & $1.28$ & $1.56$ & $7.03$ & $1.29$ & $1.86$ & $2.79$ \\
BCQ & $1.21$ & $1.38$ & $1.71$ & $1.36$ & $2.70$ & $2.15$ & $2.58$ & $2.32$ \\
BC  & $0.08$ & $0.49$ & $0.46$ & $0.39$ & $0.79$ & $0.86$ & $0.62$ & $0.83$ \\
IQL & $0.11$ & $0.43$ & $0.24$ & $0.33$ & $0.74$ & $0.87$ & $0.68$ & $0.83$ \\
\textbf{Ours}
& $2.03$ & $-0.06$ & $0.47$ & $0.51$
& $3.26$ & $3.63$ & $2.94$ & $3.49$ \\
\bottomrule
\end{tabular}
}
\end{table}

\paragraph{\bfseries\itshape Semantic-token diagnostics.}
The semantic-view variants provide the main functional diagnostic: removing any
view lowers the selected $10$-step ALL estimate under the corresponding learned
model, supporting the participation of all three semantic views in the complete
policy configuration.

\begin{figure}[t]
    \centering
    \Description{Attention heatmap showing stratum-wise relative attention of the Stage II policy on MIMIC-III across Profile, Context, Advisory, semantic, RTG, and State token groups.}
    \includegraphics[width=\columnwidth]{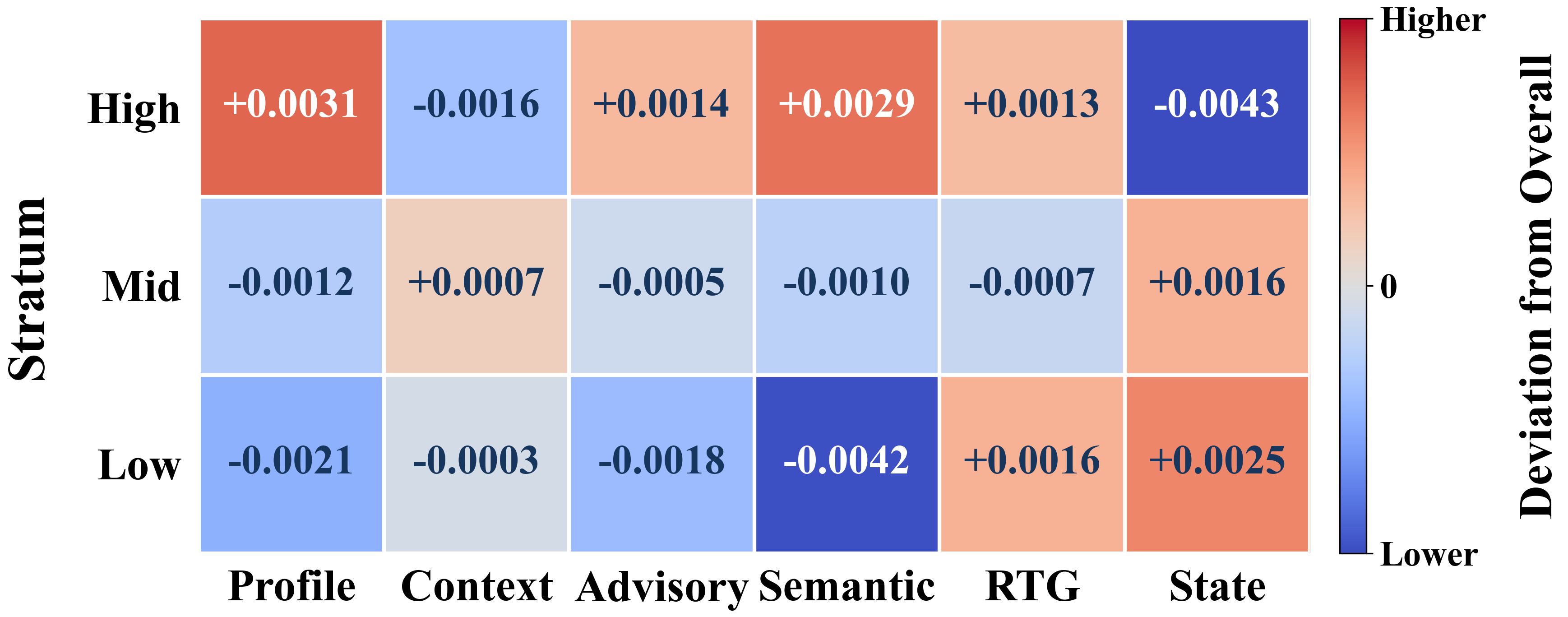}
    \caption{
    Stratum-wise relative attention heatmap of the Stage II policy on MIMIC-III.
    Warmer and cooler colors indicate higher and lower relative attention, respectively.
    }
    \label{fig:attention_deviation_mimic3}
\end{figure}

Table~\ref{tab:attention_deviation_mimic3} further reports small stratum-wise
attention deviations. HIGH-acuity trajectories place relatively more attention on
\textsc{Profile} and \textsc{Advisory}, whereas LOW-acuity trajectories shift
toward RTG and State. The ablation provides functional evidence of input
sensitivity, while attention describes how this sensitivity is organized across
acuity strata.
Figure~\ref{fig:attention_deviation_mimic3} provides a heatmap visualization of the same stratum-wise attention pattern.

\textbf{Stage II gives ClinSDT the largest $10$-step ALL change as a complete
system, and the diagnostics show sensitivity to its role-separated semantic
inputs.}

\begin{table}[t]
\centering
\scriptsize
\setlength{\tabcolsep}{3.5pt}
\renewcommand{\arraystretch}{0.90}
\caption{
Stratum-wise relative attention scores of the Stage II policy on MIMIC-III.
Positive values indicate higher relative attention to the corresponding token group, while negative values indicate lower relative attention.
}
\label{tab:attention_deviation_mimic3}
\resizebox{\columnwidth}{!}{
\begin{tabular}{lrrrrrr}
\toprule
\textbf{Stratum} & \textbf{Profile} & \textbf{Context} & \textbf{Advisory} & \textbf{Sem.} & \textbf{RTG} & \textbf{State} \\
\midrule
HIGH & $+0.0031$ & $-0.0016$ & $+0.0014$ & $+0.0029$ & $+0.0013$ & $-0.0043$ \\
MID  & $-0.0012$ & $+0.0007$ & $-0.0005$ & $-0.0010$ & $-0.0007$ & $+0.0016$ \\
LOW  & $-0.0021$ & $-0.0003$ & $-0.0018$ & $-0.0042$ & $+0.0016$ & $+0.0025$ \\
\bottomrule
\end{tabular}
}
\end{table}

\subsection{Rollout Consistency and Behavior-Support Diagnostics (RQ4)}

RQ4 examines whether ClinSDT's ALL-patient advantage persists across rollout
horizons and remains compatible with the logged action distribution. ClinSDT
retains the highest ALL-patient point estimate at both $H=5$ and $H=10$ on
MIMIC-III and MIMIC-IV, showing a consistent top-ranked trend under the
implemented protocol.

Table~\ref{tab:support_diagnostics_detail} further shows a low overall OOD ratio
of $4.84\%$ and nearly identical predicted and behavior action dispersion
($\sigma_{\text{pred}}=0.937$ vs. $\sigma_{\text{beh}}=0.938$). The OOD ratios
also remain close across HIGH, MID, and LOW strata ($4.87\%$, $4.93\%$, and
$5.17\%$, respectively). Together, these diagnostics indicate that ClinSDT's
higher ALL estimates are obtained without a broad marginal redistribution of
actions. \textbf{ClinSDT's cross-horizon top-ranked trend and close marginal
action-distribution match support the stability of its reported advantage within
the implemented rollout protocol.}

\begin{table}[t]
\centering
\scriptsize
\setlength{\tabcolsep}{3pt}
\renewcommand{\arraystretch}{0.90}
\caption{
Detailed behavior-support statistics of the Stage II policy on MIMIC-III.
}
\label{tab:support_diagnostics_detail}
\resizebox{0.98\columnwidth}{!}{
\begin{tabular}{lrrrrrr}
\toprule
\textbf{Group} & \textbf{n} & \textbf{Top1} & $\sigma_{\text{beh}}$ & $\sigma_{\text{pred}}$ & \textbf{OOD\%} & \textbf{KL} \\
\midrule
High    & 8362  & 0.458 & 0.910 & 0.907 & 4.87 & 0.662 \\
Mid     & 18646 & 0.483 & 0.943 & 0.942 & 4.93 & 0.712 \\
Low     & 1914  & 0.399 & 1.010 & 1.012 & 5.17 & 0.883 \\
\midrule
Overall & 28922 & 0.470 & 0.938 & 0.937 & 4.84 & 0.685 \\
\bottomrule
\end{tabular}
}
\end{table}
\section{Conclusion}

This work revisits offline sepsis treatment from the perspective that
retrospective ICU trajectories encode not only physiological dynamics but also
implicit clinical semantics about patient condition, treatment response, and
risk priorities. We proposed \textbf{ClinSDT}, which uses a frozen LLM to
organize these semantics into \textsc{Profile}, \textsc{Context}, and
\textsc{Advisory} tokens while leaving treatment action generation to a
Decision Transformer. A Treatment-Response Forecaster supports
uncertainty-constrained counterfactual refinement and closed-loop rollout.
Across MIMIC-III and MIMIC-IV, ClinSDT obtains the highest ALL-patient point
estimate at both $H=5$ and $H=10$. The separated-token design also leads the
evaluated complete fusion variants at ten steps on MIMIC-III, and ClinSDT's
Stage II estimates increase at both horizons. Together, these results show that
role-separated semantic abstraction can complement numerical trajectories and
provide a coherent interface between LLM-encoded clinical context and offline
sequence policy learning, with the reported advantage established within the
implemented rollout protocol.

\appendix

\section{Dataset and Evaluation Cohort Details}
\label{app:data_impl}

\subsection{Preprocessing and Trajectory Construction}

We follow the AI Clinician adult Sepsis-3 cohort pipeline. We use 4-hour bins,
LOCF/median imputation, and Z-normalization from training statistics. States
contain 45 physiological features; actions are the 25 fluid--vasopressor pairs.
We encode actions as $a=5b_{\mathrm{fluid}}+b_{\mathrm{vaso}}$, use one-step
negative SAPS-II change as reward, and compute RTG with discount $0.99$ and
scale $10$.

\begin{table}[H]
\centering
\scriptsize
\setlength{\tabcolsep}{3pt}
\renewcommand{\arraystretch}{0.90}
\caption{Preprocessing and trajectory construction.}
\label{tab:preprocessing_details}
\begin{tabular}{@{}p{0.26\columnwidth}p{0.68\columnwidth}@{}}
\toprule
\textbf{Component} & \textbf{Implementation} \\
\midrule
Cohort & Adult Sepsis-3 trajectories following the AI Clinician pipeline. \\
Temporal unit & Non-overlapping 4-hour bins. \\
State & 45 variables covering demographics, vital signs, laboratory tests,
fluid input/output, and ventilation. \\
Missing values & Last observation carried forward, followed by
population-median imputation. \\
Normalization & Z-normalization using training-set statistics. \\
Action space & $5\times5$ discretization of intravenous fluid and
vasopressor doses. \\
Action index & $a=5b_{\mathrm{fluid}}+b_{\mathrm{vaso}}$. \\
Reward / RTG & Negative one-step SAPS-II change; discount $0.99$ and scale
$10$ for RTG. \\
\bottomrule
\end{tabular}
\end{table}

\subsection{Dataset Statistics and Rollout Eligibility}

\begin{table}[H]
\centering
\scriptsize
\setlength{\tabcolsep}{3.2pt}
\renewcommand{\arraystretch}{0.92}
\caption{
Dataset statistics and rollout eligibility.
A trajectory is 10-step eligible when it contains at least 11 observations.
}
\label{tab:dataset_statistics}
\resizebox{\columnwidth}{!}{
\begin{tabular}{lcc}
\toprule
\textbf{Item} & \textbf{MIMIC-III} & \textbf{MIMIC-IV} \\
\midrule
Train trajectories & 13,011 & 52,380 \\
Validation trajectories & 2,781 & 11,225 \\
Test trajectories & 2,786 & 11,224 \\
10-step eligible test trajectories & 1,966 & 8,153 \\
Trajectory length & 2--20, mean 13.4 & 6--768, mean 26.9 \\
\bottomrule
\end{tabular}
}
\end{table}

\subsection{Acuity Stratification}

\begin{table}[H]
\centering
\scriptsize
\setlength{\tabcolsep}{3.2pt}
\renewcommand{\arraystretch}{0.92}
\caption{
Initial SAPS-II distribution and acuity stratification.
Counts are computed on the 10-step eligible test subset.
}
\label{tab:saps_stratification}
\begin{tabular}{lcc}
\toprule
\textbf{Item} & \textbf{MIMIC-III} & \textbf{MIMIC-IV} \\
\midrule
Initial SAPS-II mean & 81.8 & 53.5 \\
Initial SAPS-II range & 65.5--91.5 & 37--86 \\
Initial SAPS-II q10/q50/q90 & 79.5 / 80.5 / 91.5 & 44 / 53 / 64 \\
\midrule
\textsc{Low} threshold & $<75$ & $<50$ \\
\textsc{Mid} threshold & $[75,85)$ & $[50,57)$ \\
\textsc{High} threshold & $\geq85$ & $\geq57$ \\
\midrule
\textsc{High} count & 500 & 2,909 \\
\textsc{Mid} count & 1,328 & 2,544 \\
\textsc{Low} count & 138 & 2,700 \\
\bottomrule
\end{tabular}
\end{table}

\section{Clinical Semantic Signal Generation}
\label{app:prompts}

\subsection{Role-Separated Semantic Views}

Deterministic templates construct the views in Section~\ref{sec:semantic-construction}.
Factual views may use recorded acuity. Predicted-state views use only the
predicted state or incoming transition, except that Stage II keeps
\textsc{Profile} anchored to the factual pre-decision condition. In rollout, its
state-only template omits return-target text while the separate RTG token remains
fixed. The frozen LLM encodes descriptions but never recommends actions.

\begin{table}[H]
\centering
\scriptsize
\setlength{\tabcolsep}{2.8pt}
\renewcommand{\arraystretch}{0.90}
\caption{Semantic inputs and leakage-control regimes.}
\label{tab:semantic_input_regimes}
\resizebox{\columnwidth}{!}{%
\begin{tabular}{lll}
\toprule
\textbf{Component} & \textbf{Available input} & \textbf{Use} \\
\midrule
\textsc{Profile} & Factual / Stage II: $(s_t,m_t,R_t)$; rollout $1+$: $\hat s_k$ & Global condition \\
\textsc{Context} & Factual or generated incoming transition & Treatment response \\
\textsc{Advisory} & Factual: $(s_t,m_t)$; predicted state: $\hat s$ & Current risk priorities \\
\midrule
Acuity-enriched & Factual views; Stage II \textsc{Profile}; rollout step 0 & Recorded information only \\
Predicted-state & Stage II \textsc{Advisory}; rollout \textsc{Profile}/\textsc{Advisory} & No recorded acuity \\
\bottomrule
\end{tabular}%
}
\end{table}

\subsection{Representative Semantic Descriptions}

\begin{table}[H]
\centering
\scriptsize
\setlength{\tabcolsep}{3pt}
\renewcommand{\arraystretch}{0.92}
\caption{Representative semantic descriptions produced by the templates.}
\label{tab:semantic_examples_compact}
\begin{tabular}{@{}p{0.14\columnwidth}p{0.80\columnwidth}@{}}
\toprule
\textbf{View} & \textbf{Example description} \\
\midrule
\textsc{Profile} & Severe sepsis with cardiovascular and renal involvement; mechanically ventilated; high acuity. \\
\textsc{Context} & After vasopressor titration, blood pressure is stable, lactate is decreasing, and perfusion is improving. \\
\textsc{Advisory} & Persistent hyperlactatemia indicates high hypoperfusion risk; prioritize cardiovascular protection. \\
\bottomrule
\end{tabular}
\end{table}

\section{Implementation Details}
\label{app:hp}

\subsection{Model Architecture and Hyperparameters}

\begin{table}[H]
\centering
\scriptsize
\setlength{\tabcolsep}{3pt}
\renewcommand{\arraystretch}{0.90}
\caption{
Model architecture and main hyperparameters.
The same settings are used for MIMIC-III and MIMIC-IV unless otherwise stated.
}
\label{tab:hp}
\resizebox{0.98\columnwidth}{!}{
\begin{tabular}{ll}
\toprule
\textbf{Item} & \textbf{Value} \\
\midrule
Policy layers / heads / hidden size & $4$ / $8$ / $128$ \\
Context / tokens / block size & $20$ / $6$ / $120$ \\
LLM embedding dimension & $896$ per view \\
Forecaster type / hidden size & 3-layer MLP / $256$ \\
MC Dropout rate / samples & $0.2$ / $10$ \\
Policy / forecaster learning rate & $6\times10^{-4}$ / $10^{-3}$ \\
Batch size / weight decay & $64$ / $0.1$ \\
Stage I / II epochs; II updates per epoch & $100$ / $50$; $1000$ \\
$\rho$ / $\sigma_{\mathrm{thr}}$ & $2$ / $2.0$ \\
$\kappa$ / $\alpha_{\min}$ / $\lambda_{\Omega}$ & $0.5$ / $0.3$ / $0.5$ \\
\bottomrule
\end{tabular}
}
\end{table}

Fusion variants compress the three semantic views into one independent semantic
token. Together with the RTG, state, and action tokens, this yields four tokens
per step and a block size of $20\times4=80$.

\subsection{Two-Stage Training Procedure}
\label{app:training_eval}

Table~\ref{tab:training_eval_protocol} complements Section~3: Stage I uses
factual trajectories, whereas Stage II uses model-screened associations.

\begin{table}[H]
\centering
\scriptsize
\setlength{\tabcolsep}{3pt}
\renewcommand{\arraystretch}{0.90}
\caption{Compact training and evaluation protocol.}
\label{tab:training_eval_protocol}
\begin{tabular}{@{}p{0.21\columnwidth}p{0.73\columnwidth}@{}}
\toprule
\textbf{Stage} & \textbf{Protocol} \\
\midrule
Stage I policy & Behavior cloning on documented clinician actions. \\
Stage I forecaster & Factual state--action next-state and SAPS-II supervision. \\
Candidate set & Radius-$2$ grid neighborhood plus the policy proposal. \\
Screening & Require uncertainty below $2.0$ and lower predicted one-step SAPS-II change than the documented action. \\
Distillation & Regenerate next-step hindsight semantics and apply MSE-to-one-hot on accepted actions. \\
Stage II forecaster & Retain factual anchoring and confidence-gated bootstrap regularization; no BC or KL term. \\
Rollout & Use each method's updated forecaster; initialize DT-style RTG from the held-out trajectory and keep it fixed. \\
Metric & Negative cumulative forecaster-predicted SAPS-II change; method-specific rather than a shared clinical score. \\
\bottomrule
\end{tabular}
\end{table}

\paragraph{Stage II details.}
For $t>0$, the radius-$2$ neighborhood and policy proposal are screened using
$M=10$ MC-dropout passes; acceptance requires uncertainty below $2.0$ and a
lower one-step SAPS-II change than the documented action. For accepted $a_t^*$,
the MC-averaged $\hat{s}^{\mathrm{cf}}_{t+1}$ regenerates next-step
\textsc{Context} and \textsc{Advisory}, while \textsc{Profile} remains tied to
$s_t$. These training-only semantics supervise MSE-to-one-hot distillation.
Stage II retains factual anchoring and confidence-gated bootstrap regularization
but no BC or KL term; rollout instead refreshes semantics from generated states
and incoming transitions.

\section{Evaluation Protocol and Reporting Scope}
\label{app:evaluation_scope}

\subsection{Closed-Loop Rollout and Scoring}
\label{app:rollout_scoring}

Each method uses its own Stage-II-updated forecaster. DT-style methods keep the
held-out initial RTG fixed; non-DT baselines use none. The metric
$-\sum_{t=0}^{H-1}\bar d^{\mathrm{saps}}_{\phi_i,t}$ compares complete systems
under a common retrospective procedure, not a shared clinical value scale.

\subsection{Information Boundaries}
\label{app:information_boundaries}

Table~\ref{tab:information_boundaries} separates Stage II's training-only
counterfactual predictions from rollout, which cannot use recorded future states.

\begin{table}[H]
\centering
\scriptsize
\setlength{\tabcolsep}{3pt}
\renewcommand{\arraystretch}{0.90}
\caption{Information available across training and evaluation phases.}
\label{tab:information_boundaries}
\begin{tabular}{@{}p{0.23\columnwidth}p{0.71\columnwidth}@{}}
\toprule
\textbf{Phase} & \textbf{Available information} \\
\midrule
Stage I semantics & Factual current/incoming transition and, where specified,
recorded acuity. \\
Stage II screening & Current factual state, documented action, local
candidates, and forecaster predictions. \\
Stage II regeneration & Current factual state, accepted candidate, and its
predicted counterfactual next state; training only. \\
Rollout step 0 & Held-out initial state and recorded initial information; no
subsequent observation. \\
Rollout steps $1+$ & Previously generated state and incoming generated
transition; state-only semantic regime. \\
Scoring & Predicted SAPS-II changes from the corresponding method-specific
updated forecaster. \\
\bottomrule
\end{tabular}
\end{table}

\subsection{Score Aggregation and Interpretation}

For a given dataset and horizon, all methods start from the same eligible test
trajectories. Acuity strata are assigned once from the initial SAPS-II score
and remain fixed throughout rollout. The \textsc{All} result pools the eligible
trajectories directly rather than averaging the three stratum-level scores.
Because each policy advances its generated state with its own learned
forecaster, a cross-method difference may reflect the policy, the forecaster
dynamics, or their interaction. The tables therefore compare complete systems
within the same dataset and horizon, not causal treatment effects or calibrated
units of clinical benefit.

This phase separation keeps recorded future states out of closed-loop rollout
inputs, but neither removes forecaster error nor makes generated trajectories
equivalent to observed clinical outcomes. The results therefore support only
protocol-relative assessment of the complete systems; prospective utility and
independent off-policy validation remain outside the present scope. The same
reporting rules are applied to every method.

\begin{acks}
The work described in this paper was partially supported by InnoHK initiative,
The Government of the HKSAR, and Laboratory for AI-Powered Financial
Technologies.
\end{acks}

\clearpage

\section*{GenAI Usage Disclosure}

Generative AI assisted only with language polishing; the authors remain
responsible for all technical content, experiments, analysis, and conclusions.
It was not used to generate results, data, references, or treatment decisions.
The frozen LLM studied in ClinSDT is part of the method and serves only as a
semantic encoder; the policy model generates all treatment actions.

\bibliographystyle{ACM-Reference-Format}
\balance
\bibliography{references}

\end{document}